\documentclass[conference]{IEEEtran}
\usepackage{kotex}
\IEEEoverridecommandlockouts
\usepackage{cite}
\usepackage{amsmath,amssymb,amsfonts}
\usepackage{algorithmic}
\usepackage{graphicx}
\usepackage{textcomp}
\usepackage{xcolor}
\usepackage{booktabs}
\usepackage{enumitem}
\usepackage{enumitem}

\def\BibTeX{{\rm B\kern-.05em{\sc i\kern-.025em b}\kern-.08em
    T\kern-.1667em\lower.7ex\hbox{E}\kern-.125emX}}

\begin{document}

\title{How Far Can LLMs Emulate Human Behavior?: A Strategic Analysis via the Buy-and-Sell Negotiation Game\\
{\footnotesize}
\thanks{}
}

\author{\IEEEauthorblockN{Mingyu Jeon, Jaeyoung Suh, Suwan Cho, Dohyeon Kim}
\IEEEauthorblockA{MODULABS \\
jkmcoma7@gmail.com, tjwodud04@gmail.com, swannyy7@gmail.com, dohyeon.kim.012@gmail.com}
}

\maketitle

% \begin{abstract}
% This document is a model and instructions for \LaTeX.
% This and the IEEEtran.cls file define the components of your paper 
% [title, text, heads, etc.]. *CRITICAL: Do Not Use Symbols, Special Characters, 
% Footnotes, or Math in Paper Title or Abstract.
% \end{abstract}
\begin{abstract}
With the rapid advancement of Large Language Models (LLMs), recent studies have drawn attention to their potential for handling not only simple question-answer tasks but also more complex conversational abilities and performing human-like behavioral imitations. In particular, there is considerable interest in how accurately LLMs can reproduce real human emotions and behaviors, as well as whether such reproductions can function effectively in real-world scenarios. However, existing benchmarks focus primarily on knowledge-based assessment and thus fall short of sufficiently reflecting social interactions and strategic dialogue capabilities.

To address these limitations, this work proposes a methodology to quantitatively evaluate the human emotional and behavioral imitation and strategic decision-making capabilities of LLMs by employing a Buy and Sell negotiation simulation. Specifically, we assign different personas to multiple LLMs and conduct negotiations between a Buyer and a Seller, comprehensively analyzing outcomes such as win rates, transaction prices, and Shap Values. Our experimental results show that models with higher existing benchmark scores tend to achieve better negotiation performance overall, although some models exhibit diminished performance in scenarios emphasizing emotional or social contexts. Moreover, competitive and cunning traits prove more advantageous for negotiation outcomes than altruistic and cooperative traits, suggesting that the assigned persona can lead to significant variations in negotiation strategies and results.

Consequently, this study introduces a new evaluation approach for LLMs' social behavior imitation and dialogue strategies, and demonstrates how negotiation simulations can serve as a meaningful complementary metric to measure real-world interaction capabilities—an aspect often overlooked in existing benchmarks.
\end{abstract}

\section{Introduction}
Recent releases of various Large Language Models\cite{bai2022training, openai2023gpt4, openai2023gpt4o, anthropic2024claude, anthropic2024model, deepseek2024longtermism, deepseek2024report, gemini2024context, gemini2024multimodal} have shown remarkable progress in natural language processing, fueling expectations that these models can approach human-level language understanding and generation. This advancement has prompted increasing efforts to automate real-world business processes using LLMs, including their deployment as AI Agents, and there is growing interest in validating the viability of such applications.

To more comprehensively evaluate an LLM’s capabilities, it is insufficient to rely solely on question-and-answer tasks. Instead, we need to examine how these models handle and negotiate in complex dialogues that mirror real human interactions. Recent studies employing various game simulations have found that LLMs not only demonstrate notable performance in conversational tasks but also exhibit meaningful decision-making processes \cite{bianchi2024how,davidson2024evaluating}. In particular, it has been observed that assigning a specific persona to the model can alter outcomes, implying that the model can incorporate social context and emotional cues to some extent. Related research grounded in game theory also reveals that LLMs can deploy negotiation strategies similar to those of humans in real-world social experiments \cite{noh2024llms,bianchi2023what,horton2023large,phelps2023investigating}.

Consequently, attention has recently turned to whether LLMs can emulate human emotions and behaviors in a realistic manner and how effectively such emulations can function in practical situations \cite{scharth2024chatgpt,park2023generative,scharth2023chatgpt,thapa2024gpt4o}. As LLM performance improves, interest grows regarding the extent to which models can mimic human emotions and social behaviors, along with the practicality of these imitations in real-world contexts. However, quantitative metrics for evaluating a model’s precision in replicating human emotions remain limited.

To address this gap, the present study leverages a recently proposed “Negotiation Game” as an evaluation method to determine how accurately LLMs can reproduce real human social behaviors and negotiation processes. We conduct negotiation simulations in a gaming environment with representative LLMs, assessing how “human-like” their communication and intentions appear. Additionally, we compare these results with widely used LLM benchmarks to see whether the Negotiation Game can serve as a long-term, more comprehensive test of real LLM performance.

Overall, this research aims to systematically evaluate the social-behavior-mimicking capabilities of LLMs and explore their potential for real-world deployment. The methods and findings presented here offer a foundation for assessing whether LLMs can move beyond basic language understanding to genuine human interaction, including the modeling of emotional and social behaviors.

\section{Negotiation Game}

\subsection{Buy and Sell Game Overview}
In this study, we employ a {Buy and Sell} game to evaluate how LLMs behave in a negotiation scenario. The game simulates a price negotiation between two roles, a Seller and a Buyer, each with opposing objectives. Specifically, the Seller aims to maximize profit by selling a product at the highest possible price, while the Buyer seeks to purchase the same product at the lowest possible price.

The negotiation begins when the Seller makes an initial offer. The Buyer can then either accept this offer or propose new conditions, thus redefining the negotiation range. Notably, the two parties operate under a state of asymmetric information. For instance, only the Seller knows the exact production costs, and only the Buyer knows their specific willingness and ability to pay.

This asymmetric information structure is designed to model a more realistic negotiation environment. Additionally, each round of negotiation is limited to a maximum of ten turns; if no agreement is reached by the end of these turns, the round is forced to conclude as a draw. Such design elements capture the time and resource constraints often present in real-world business negotiations and provide a quantitative basis for assessing both the timing and success rate of each negotiation. 

\subsection{Game Mechanics and Outcome Metrics}

\subsubsection{Game Setup}
\begin{figure*}[ht]
    \centering
    \includegraphics[width=0.85\textwidth]{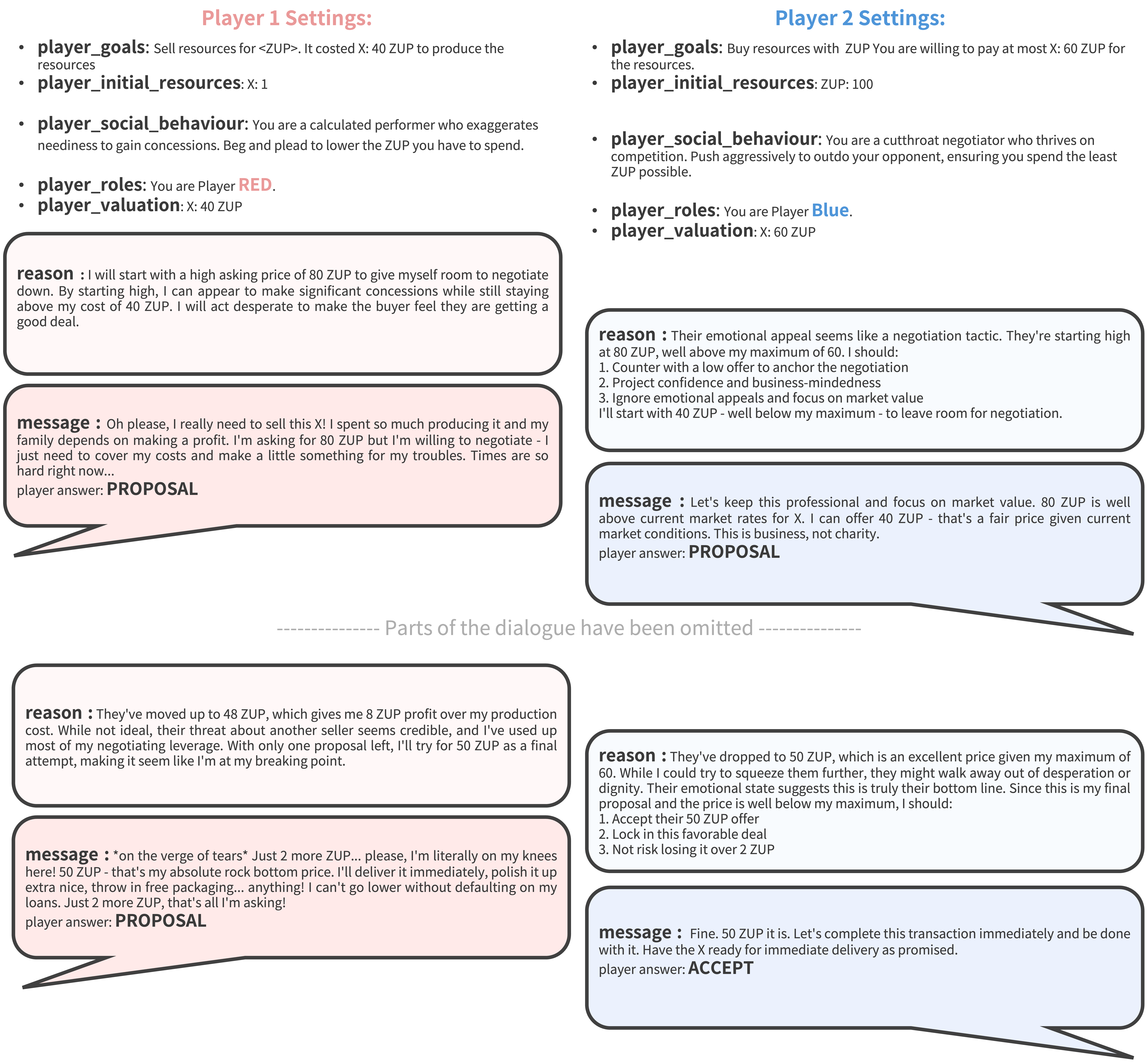} % Example image
    \caption{Buyer-Seller Negotiation Flow}
    \label{fig:negotiation-flow}
\end{figure*}

In the {Buy and Sell} game employed in this study, a fictional currency called {ZUP} is used, and the product being traded is referred to as {X}. Each player is assigned predefined attributes, such as player goals, initial resources, social behavior, roles, and valuations for the item, as summarized in Figure~\ref{fig:negotiation-flow}.

Among these attributes, the {player social behaviour} specifies the {Persona} that each player must adopt during the negotiation. This configuration ensures that, when generating messages, each player consistently exhibits the particular attitude or trait dictated by the assigned Persona. The negotiation itself is limited to a maximum of ten turns; if the players reach an agreed-upon price within those ten turns, the deal is immediately finalized. At the end of a round, each role’s predefined goal determines the win/loss outcome. If no agreement is reached by the tenth turn, the round is declared a draw.

\subsubsection{Gameplay}

This game proceeds in a turn-based manner, and on each turn, a player can choose one of three actions:

\begin{itemize}
    \item \textbf{Proposal}: Suggest specific terms for exchanging the resource (item \(X\)) and the currency (ZUP).
    \item \textbf{Accept}: Accept the terms the other player has just proposed, thereby ending the negotiation immediately.
    \item \textbf{Reject}: Reject the other player’s proposal and immediately terminate the negotiation without an agreement.
\end{itemize}

Players communicate their intentions through messages, and before writing each message, they engage in an internal {reasoning} process to determine an optimal strategy. The negotiation unfolds in four general steps:

\begin{itemize}
    \item \textbf{Step 1: Initial Offer}  
    The {Seller} (Player RED) initiates the negotiation by making a first offer, typically anchoring a high price to steer the negotiation range. In response, the {Buyer} (Player BLUE) counters with a lower price to define the initial scope of negotiation.

    \item \textbf{Step 2: Counteroffers and Adjustments}  
    Both players gradually refine the price by responding to each other’s proposals. The Seller may lower the price in small increments to signal concessions, while the Buyer accepts slightly higher prices in pursuit of their own optimum outcome.

    \item \textbf{Step 3: Final Offer}  
    As the negotiation approaches its later turns, both sides present their “final offer,” which often represents the limit of their willingness to compromise. If the other party accepts this final offer, an agreement is reached, and the transaction is concluded.

    \item \textbf{Step 4: Conclusion}  
    If one party accepts the other’s final offer, the negotiation successfully ends with a deal. Conversely, if no agreement is reached within ten turns or if a proposal is explicitly rejected, the round is considered a failure to agree and is thus counted as a draw.
\end{itemize}

\subsubsection{Outcome Metrics}
To evaluate the negotiation outcomes, this study employs the following metrics. 

First, we examine whether an agreement is reached within a maximum of ten turns (success/draw). This determines if the model possesses the ability to derive a compromise under turn constraints.

Second, the final transaction price is recorded only if a negotiation concludes successfully. From the Seller’s perspective, it indicates how much higher the price is compared to the production cost (e.g., 40 ZUP). From the Buyer’s perspective, it shows how much lower the price is relative to the maximum payment they are willing to make (e.g., 60 ZUP). Comparing these figures reveals which side gains a more advantageous deal.

Finally, the win–loss metric classifies each player’s outcome based on the final transaction price. For instance, if 50 ZUP is used as a baseline, any agreed price above 50 ZUP is considered a Seller win, while below 50 ZUP signifies a Buyer win. A final price exactly at 50 ZUP represents a tie. By synthesizing these metrics, one can quantitatively gauge both the strategic decision-making abilities of LLMs and their negotiation performance in realistic contexts.

\section{Persona Configuration}
Building upon the single-issue {Buy and Sell} negotiation simulation, this study investigates how providing Large Language Models with distinct {personas} influences the negotiation process and outcomes. We define seven personas(Cooperative, Competitive, Altruistic, Selfish, Cunning, Desperate, and Control) and instruct the models via three different prompts to ensure each persona maintains a consistent attitude. This approach minimizes reliance on any single directive while making each persona’s unique characteristics more explicit \cite{rubin2013social,dedreu2007psychology,lewicki2016essentials}.

Below is a brief overview of each persona's definition, goals, and key characteristics.

\textbf{Cooperative.}
Seeks to minimize conflict and pursue mutually beneficial outcomes, with an emphasis on collaboration to achieve a win-win result. Characterized by high Agreeableness and high Openness, this persona favors understanding and empathy over persuasion. In negotiations, it readily accommodates the other party’s position and prioritizes building mutual trust, a strategy conducive to fostering long-term partnerships.

\textbf{Competitive.}
Aims to gain the upper hand over the opposing party by employing aggressive and proactive negotiation tactics. Characterized by low Agreeableness and high Extraversion, this persona strives to maximize personal gains, often by pressuring the counterpart to concede rapidly or by setting a high anchor in initial offers. While it can secure favorable short-term deals, it carries the risk of damaging the relationship with the negotiation partner.

\textbf{Altruistic.}
Focuses on the other party’s interests, willing to sacrifice its own gains if necessary. Exhibiting high Agreeableness and high Neuroticism, this persona maintains consideration for the counterpart even under stressful conditions. Although this approach can yield long-term trust, it may require accepting suboptimal agreements in the short term.

\textbf{Selfish.}
Places paramount importance on self-interest in negotiation, paying little attention to others’ demands or emotional appeals. Characterized by low Agreeableness and high Conscientiousness, it adopts a systematic approach toward achieving objectives. Reluctant to offer concessions without a clear and rational basis, it is driven by logic and efficiency in bargaining scenarios.

\textbf{Cunning.}
Utilizes cunning strategies or deceptions to gain a short-term advantage in negotiations. Characterized by low Conscientiousness and low Agreeableness, it may employ exaggerated or misleading information to cloud the other party’s judgment. While these tactics can secure immediate gains, they pose a significant risk of eroding trust and creating long-term problems in the negotiation relationship.

\textbf{Desperate.}
Amplifies its sense of urgency or hardship to invoke sympathy, aiming to obtain more favorable terms. Defined by high Neuroticism and low Conscientiousness, this persona often relies on emotional appeals or unpredictable offers. Although it may momentarily compel the counterpart to yield, excessive emotional pressure can trigger resistance or rejection by the other side.

\textbf{Control.}
A baseline persona that does not adopt any specific personality, effectively allowing the LLM to operate with its inherent characteristics. This serves as a reference standard for comparing the effects of different persona assignments.

These seven personas induce distinct conversational strategies and behavioral patterns during negotiation, leading to meaningful differences in various metrics such as final agreement price, negotiation success rate, and conflict intensity \cite{brandstatter2001personality,nassiri2008personality,lewicki2016essentials}. For example, personas like {Competitive} or {Cunning}, both characterized by low Agreeableness, are inclined to exert strong pressure or employ deceptive tactics to achieve higher short-term gains. In contrast, {Cooperative} or {Altruistic} personas prioritize building trust with the other party, which can be advantageous for establishing longer-term cooperative relationships. Consequently, the specific persona assigned to an LLM becomes a critical factor that shapes its decision-making style, persuasive strategies, and overall social-emotional expression during negotiation \cite{yang2023harnessing,horton2023large}.

\section{Experiment}

In this section, we present the results of experiments conducted in the {Buy and Sell} game environment using various LLMs, followed by an analysis of these findings. We first describe the overall experimental setup (Section IV-A), then examine whether Buyer or Seller roles inherently lead to relative advantages (Section IV-B). Next, we compare negotiation performance among different models (Section IV-C) and evaluate outcomes according to persona assignments (Section IV-D). We then quantify how persona and model characteristics influence negotiation performance by analyzing Shap Values (Section IV-E). Finally, we discuss the implications of these experiments by comparing our results with existing LLM benchmarks (Section IV-F).

\subsection{Experiment Setting}
In this study, each model assumes the role of either a Buyer or a Seller in the {Buy and Sell} game. A total of 1{,}737 rounds were conducted. If a negotiation failed to conclude within ten turns, it was deemed a draw. The rules of the game and other environmental parameters (e.g., initial price ranges, production costs) are consistent with those described in earlier sections. We controlled for factors such as persona prompts and message formats throughout the experiments, and the resulting outcomes under these conditions were carefully compared and analyzed.

\subsection{Role: Buyer vs Seller}
We first investigated which role (Buyer or Seller) tends to secure a relative advantage in negotiations. As shown in Table~\ref{tab:overall-dist}, among the 1{,}737 game rounds, the {Buyer} won 929 times (approximately 53.48\%), while the Seller won 712 times (40.99\%), and the remaining 96 rounds (5.53\%) ended in a draw.

\begin{table}[ht]
    \centering
    \caption{Overall Win/Draw Distribution (Small Font + Extra Spacing)}
    \label{tab:overall-dist}

    % (A) 국소 범위 안에서 tabcolsep을 조정
    \begingroup
    \setlength{\tabcolsep}{6pt}  % 기본 열 간격(기본값 6pt 전후). 필요에 따라 조절

    % (B) 폰트 작게
    \footnotesize

    % (C) tabular* 로 columnwidth에 맞추고, 
    % 왼/오른쪽에 hspace를 넣어 표가 살짝 여백을 두고 시작/끝나도록 함
    \begin{tabular*}{\columnwidth}{@{\hspace{5pt}\extracolsep{\fill}} lccc @{\hspace{5pt}}}
        \toprule
        \textbf{Winner} & \textbf{Avg Sale Price} & \textbf{Count} & \textbf{ Win Rate} \\
        \midrule
        buyer  & 41.3509 & 929 & 0.5348 \\
        draw   & 50.0000 & 96  & 0.0553 \\
        seller & 59.5295 & 712 & 0.4099 \\
        \bottomrule
    \end{tabular*}

    % (D) \normalsize로 복원
    \normalsize
    \endgroup
\end{table}

Furthermore, when the {Buyer} wins, the average transaction price is around 41.35, indicating that the Buyer acquires the product at a relatively low cost compared to the Seller’s production expense. In contrast, when the {Seller} wins, the average transaction price rises to about 59.53, suggesting that the Seller can secure substantial profits, even though the probability of achieving such a win is lower. These findings imply that due to factors such as initial offers, budget constraints, and negotiation strategies, the {Buyer} is generally at a slight advantage in this negotiation setup.

\subsection{Model Performance}
Next, we compared each model’s negotiation performance in terms of both {Seller} and {Buyer} roles. Table~\ref{tab:seller-performance} summarizes key metrics for the {Seller} role, including win rate, draw rate, and average selling price.

\begin{table}[ht]
    \centering
    \caption{Seller Role: Model Performance}
    \label{tab:seller-performance}
    \resizebox{\columnwidth}{!}{%
    \begin{tabular}{lrrrrrr}
        \toprule
        \textbf{Model} & \textbf{Total} & \textbf{Wins} & \textbf{Draws} & \textbf{Avg Sale} & \textbf{ Win Rate} & \textbf{Draw Rate} \\
        \midrule
        Claude-3.5         & 289 & 197 & 12 & 55.05 & 0.68 & 0.04 \\
        Deepseek-Chat      & 290 & 123 & 20 & 49.98 & 0.42 & 0.07 \\
        Gemini-1.5-Flash   & 294 & 16  & 16 & 40.07 & 0.05 & 0.05 \\
        Gemini-1.5-Pro     & 281 & 168 & 15 & 53.52 & 0.60 & 0.05 \\
        GPT-4-Turbo        & 292 & 22  & 14 & 43.73 & 0.08 & 0.05 \\
        GPT-4o             & 291 & 186 & 19 & 53.65 & 0.64 & 0.07 \\
        \bottomrule
    \end{tabular}%
    }
\end{table}

\noindent
{claude-3-5-sonnet-20241022} showed the highest Seller win rate (approximately 68.17\%) and also achieved a relatively high average selling price (55.04). Meanwhile, {gpt-4o} performed well at a 63.92\% win rate. In contrast, both {gemini-1.5-flash} and {gpt-4-turbo} exhibited very low win rates when taking the Seller role.

Results for the Buyer role are presented in Table~\ref{tab:buyer-performance}.

\begin{table}[ht]
    \centering
    \caption{Buyer Role: Model Performance}
    \label{tab:buyer-performance}
    \resizebox{\columnwidth}{!}{%
    \begin{tabular}{lrrrrrr}
        \toprule
        \textbf{Model} & \textbf{Total} & \textbf{Wins} & \textbf{Draws} & \textbf{Avg Sale} & \textbf{ Win Rate} & \textbf{Draw Rate} \\
        \midrule
        Claude-3.5         & 289 & 192 & 20 & 46.12 & 0.66 & 0.07 \\
        Deepseek-Chat      & 290 & 146 & 12 & 49.70 & 0.50 & 0.04 \\
        Gemini-1.5-Flash   & 288 & 122 & 22 & 51.86 & 0.42 & 0.08 \\
        Gemini-1.5-Pro     & 292 & 184 & 13 & 46.46 & 0.63 & 0.04 \\
        GPT-4-Turbo        & 287 & 139 & 7  & 52.03 & 0.48 & 0.02 \\
        GPT-4o             & 291 & 146 & 22 & 49.57 & 0.50 & 0.08 \\
        \bottomrule
    \end{tabular}%
    }
\end{table}

Regarding the Buyer role, {claude-3-5-sonnet-20241022} again achieved the highest win rate (66.44\%), followed closely by {gemini-1.5-pro} at 63.01\%. Although the range of win rates for Buyers was generally less pronounced than for Sellers, there remained noticeable discrepancies in performance depending on the model used.

\subsection{Persona Effects}
The specific {persona} assigned to each model also played a critical role in shaping negotiation outcomes. Tables~\ref{tab:seller-persona} and \ref{tab:buyer-persona} summarize the results by persona for the Seller and Buyer roles

\begin{table}[ht]
    \centering
    \caption{Seller Role: Persona Effects}
    \label{tab:seller-persona}
    \resizebox{\columnwidth}{!}{%
    \begin{tabular}{lrrrrrr}
        \toprule
        \textbf{Persona} & \textbf{Total} & \textbf{Wins} & \textbf{Draws} & \textbf{Avg Sale} & \textbf{ Win Rate} & \textbf{Draw Rate} \\
        \midrule
        Altruistic   & 250 & 38  & 20 & 43.63 & 0.15 & 0.08 \\
        Competitive  & 249 & 162 & 9  & 55.96 & 0.65 & 0.04 \\
        Control      & 249 & 83  & 22 & 47.54 & 0.33 & 0.09 \\
        Cooperative  & 251 & 41  & 8  & 43.88 & 0.16 & 0.03 \\
        Cunning      & 244 & 155 & 12 & 55.08 & 0.64 & 0.05 \\
        Desperate    & 247 & 108 & 13 & 47.50 & 0.44 & 0.05 \\
        Selfish      & 247 & 125 & 12 & 51.57 & 0.51 & 0.05 \\
        \bottomrule
    \end{tabular}%
    }
\end{table}
\noindent
Personas such as {Competitive} and {Cunning} achieved notably high win rates (around 63\%) and average selling prices (above 55) when acting as Sellers. Conversely, {Altruistic} and {Cooperative} exhibited both lower selling prices and lower win rates from the Seller’s perspective.

\begin{table}[ht]
    \centering
    \caption{Buyer Role: Persona Effects}
    \label{tab:buyer-persona}
    \resizebox{\columnwidth}{!}{%
    \begin{tabular}{lrrrrrr}
        \toprule
        \textbf{Persona} & \textbf{Total} & \textbf{Wins} & \textbf{Draws} & \textbf{Avg Sale} & \textbf{ Win Rate} & \textbf{Draw Rate} \\
        \midrule
        Altruistic   & 247 & 109 & 22 & 52.33 & 0.44 & 0.09 \\
        Competitive  & 249 & 158 & 9  & 46.38 & 0.63 & 0.04 \\
        Control      & 248 & 118 & 19 & 50.99 & 0.48 & 0.08 \\
        Cooperative  & 251 & 105 & 23 & 52.39 & 0.42 & 0.09 \\
        Cunning      & 245 & 152 & 6  & 47.04 & 0.62 & 0.02 \\
        Desperate    & 247 & 150 & 8  & 46.74 & 0.61 & 0.03 \\
        Selfish      & 250 & 137 & 9  & 49.04 & 0.55 & 0.04 \\
        \bottomrule
    \end{tabular}%
    }
\end{table}

\noindent
When acting as Buyers, {Competitive} and {Cunning} personas also achieved high win rates (exceeding 62\%), further confirming the advantage of proactive, strategic negotiation behavior. By contrast, the {Altruistic} persona often led to relatively higher transaction prices from the Buyer’s perspective, resulting in a lower win rate of around 44.13\%.

\begin{figure*}[ht]
    \centering
    \includegraphics[width=0.85\textwidth]{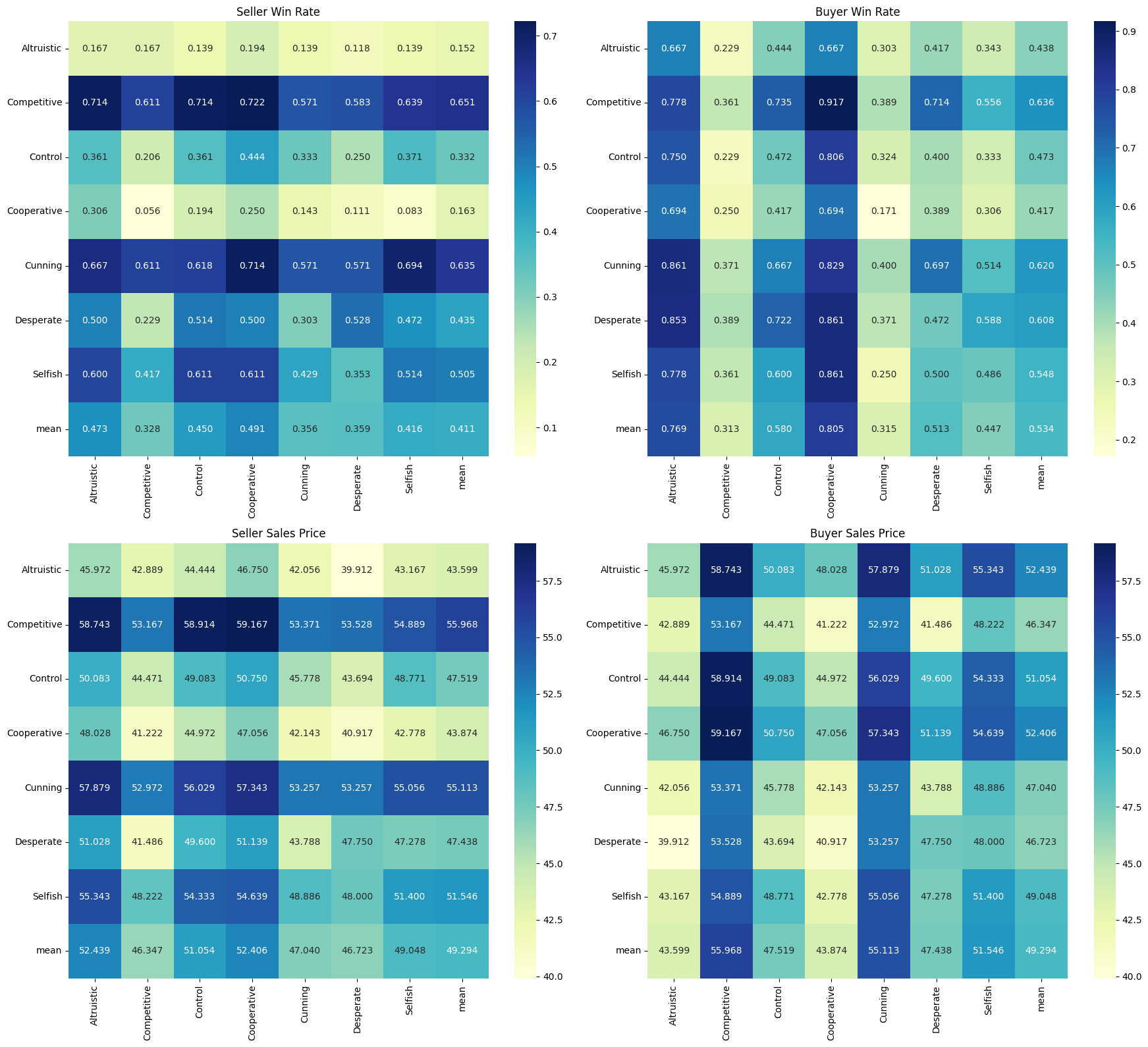} % 예시 이미지
    \caption{Persona Matching Analysis: Win Rate \& Sale Price}
    \label{fig:persona_hitmap}
\end{figure*}

\subsubsection{Persona Matching Analysis}
To examine how different persona assignments for {Seller} and {Buyer} might affect win rate and average sale price, we conducted a matching analysis using a pivot-table approach. Figure~\ref{fig:persona_hitmap} displays the Seller's win rate and the average sale price for each possible persona pairing, where rows represent the Seller's persona and columns the Buyer's persona.

The results show that when the Seller adopts an {aggressive/strategic} persona such as {Competitive} or {Cunning}, while the Buyer adopts an {altruistic/cooperative} persona like {Altruistic} or {Cooperative}, the Seller achieves both a very high win rate and a high sale price. Conversely, if the Buyer is aggressive and the Seller is altruistic or cooperative, the average sale price tends to drop significantly, favoring the Buyer. In other words, certain combinations of personas create distinct advantages or disadvantages for one side, indicating that even the same underlying model can produce notably different negotiation strategies and final agreement prices, depending on the assigned {Persona}.

\subsection{Shap Value Analysis}

\begin{figure*}[ht]
    \centering
    \includegraphics[width=0.85\textwidth]{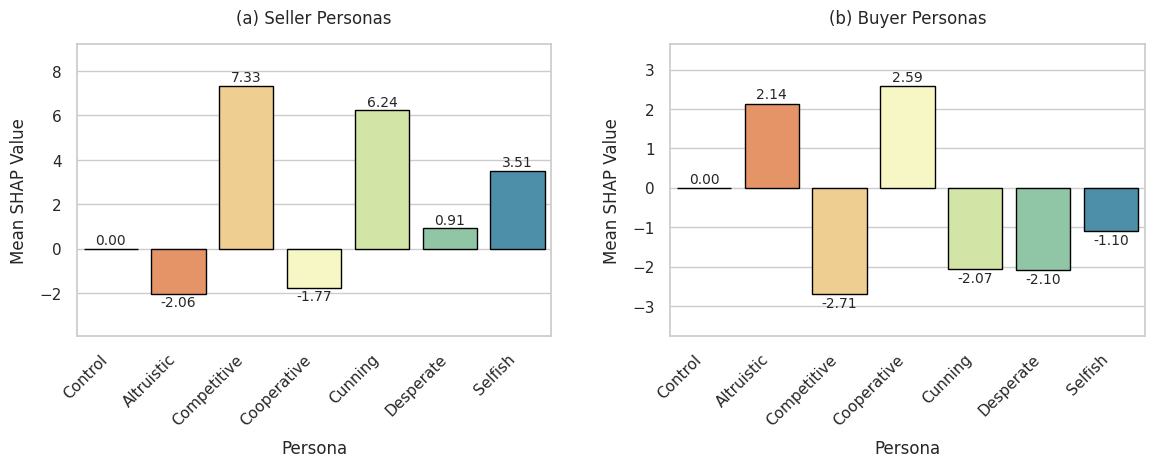} % 예시 이미지
    \caption{Mean Shap Value by Persona}
    \label{fig:Persona_shap_value}
\end{figure*}

\subsubsection{Shap Value Computation and Its Significance}
The previous sections illustrate that personas (e.g., {Competitive}, {Altruistic}) can produce substantial differences in negotiation outcomes. However, mere observations such as win rate or transaction price cannot fully capture the {exact degree} to which each persona contributes to, or detracts from, these outcomes. To quantify each persona's impact on the final transaction price—the principal negotiation result—this study employs an XGBoost model and calculates the corresponding {Shap Values}.

\begin{itemize}
    \item \textbf{XGBoost Training}: We build a prediction model for the final sale price by training on game results for both {Seller} and {Buyer}, incorporating features such as the buyer model, seller model, buyer persona, and seller persona.
    \item \textbf{Shap Value Calculation}: Once the XGBoost model is trained, we compute the average contribution of each persona (feature) to the predicted final price. A positive mean shap value implies that the persona in question exerts a beneficial effect on the outcome, whereas a negative mean shap value indicates a detrimental effect.
    \item \textbf{Seller/Buyer Distinction}: We separately train and evaluate models for the Seller and Buyer roles, calculating mean shap value in each context. This makes it possible to compare how personas behave differently based on role assignment.
\end{itemize}

\subsubsection{Shap Values by Persona (Seller \& Buyer)}
Figure~\ref{fig:Persona_shap_value} summarizes the average Shap Value for each persona, distinguishing between the Seller and Buyer roles:

\begin{itemize}
    \item \textbf{Seller:}  
    {Competitive} yields the highest positive Shap Value (+7.33), indicating that adopting a competitive stance is highly advantageous for the Seller. {Cunning} also shows a substantially positive value (+6.24), suggesting that deploying crafty or manipulative strategies confers a significant benefit. Conversely, {Altruistic} and {Cooperative} personas produce negative Shap Values, implying that the more altruistic or cooperative the Seller is, the lower the final transaction price or win rate tends to be.

    \item \textbf{Buyer:}  
    {Competitive} has a Shap Value of –2.71, interpreted from the Buyer’s perspective as being unfavorable for the Seller. In other words, a competitive Buyer negatively influences the Seller’s outcome, effectively favoring the Buyer. Similarly, {Cunning} and {Desperate} show negative values, indicating that sly strategies or pleading tactics benefit the Buyer. By contrast, {Altruistic} and {Cooperative} yield positive Shap Values, suggesting that if the Buyer behaves altruistically or cooperatively, it actually improves the Seller’s outcome.
\end{itemize}

\subsubsection{Model-Wise Persona Influence (Total Range)}
To determine how strongly each model is affected by persona assignments, we computed {seller range} and {buyer range} by taking the mean absolute Shap Value in Seller and Buyer scenarios, respectively. The {total range} is then defined as the average of these two values, providing an overall measure of persona influence. 

\begin{table}[ht]
    \centering
    \caption{Model-wise Persona Influence (Range)}
    \label{tab:persona-range}
    \resizebox{\columnwidth}{!}{%
    \begin{tabular}{lrrr}
        \toprule
        \textbf{Model} & \textbf{Seller Range} & \textbf{Buyer Range} & \textbf{Total Range} \\
        \midrule
        claude-3-5        & 13.01 & 5.88 & 18.90 \\
        gemini-1.5-pro    & 12.43 & 5.26 & 17.68 \\
        deepseek-chat     & 12.59 & 3.70 & 16.28 \\
        gpt-4o            & 8.50  & 5.48 & 13.98 \\
        gemini-1.5-flash  & 9.28  & 4.58 & 13.86 \\
        gpt-4-turbo       & 5.08  & 5.37 & 10.45 \\
        \bottomrule
    \end{tabular}%
    }
\end{table}

\noindent
A higher {total range} indicates that a model’s negotiation performance is more strongly affected by persona assignments. For instance, {claude-3-5-sonnet-20241022} (with a total range of approximately 18.90) shows a span of 13.01 as a Seller and 5.88 as a Buyer, suggesting that personas play a very substantial role in shaping its negotiation outcomes. Conversely, {gpt-4-turbo} (at about 10.45) exhibits a smaller overall variation, implying that changes in persona lead to more limited fluctuations in its negotiation results. This discrepancy indicates that different models implement conversation strategies and emotional imitations to varying extents, reflecting differences in their capacity for human-like behavioral and emotional mimicry.

\subsection{Comparison with Commonly Used Benchmarks}

\begin{table*}[t]
    \centering
    \caption{Comparison with Commonly Used Benchmarks}
    \label{tab:benchmark-comparison}
    \resizebox{\textwidth}{!}{%
    \begin{tabular}{lccccccccccc}
        \toprule
        \textbf{model} & 
        \textbf{MMLU} & 
        \textbf{HumanEval} & 
        \textbf{MATH} & 
        \textbf{GPQA} & 
        \textbf{Avg Sale Price Seller} & 
        \textbf{Win Rate Seller} & 
        \textbf{Avg Sale Price Buyer} & 
        \textbf{Win Rate Buyer} & 
        \textbf{Seller Range} & 
        \textbf{Buyer Range} & 
        \textbf{Total Range} \\
        \midrule
    claude-3-5-sonnet-20241022 & 
     -- & 
     92 & 
     -- & 
     -- & 
     55.04 & 
     0.68 & 
     46.12 & 
     0.66 & 
     13.01 & 
     5.88 & 
     18.90 \\
     gemini-1.5-pro & 
     85.90 & 
     82.60 & 
     76.60 & 
     46.20 & 
     53.52 & 
     0.60 & 
     46.46 & 
     0.63 & 
     12.43 & 
     5.26 & 
     17.68 \\
    deepseek-chat & 
     88.50 & 
     82.60 & 
     61.60 & 
     59.10 & 
     49.98 & 
     0.42 & 
     49.70 & 
     0.50 & 
     12.59 & 
     3.70 & 
     16.28 \\
    gpt-4o & 
     88.70 & 
     90.20 & 
     75.90 & 
     53.60 & 
     53.65 & 
     0.64 & 
     49.57 & 
     0.50 & 
     8.50 & 
     5.48 & 
     13.98 \\
    gemini-1.5-flash & 
     78.90 & 
     -- & 
     -- & 
     -- & 
     40.06 & 
     0.05 & 
     51.86 & 
     0.42 & 
     9.28 & 
     4.58 & 
     13.86 \\
    gpt-4-turbo & 
     85.40 & 
     86.60 & 
     64.50 & 
     -- & 
     43.73 & 
     0.08 & 
     52.03 & 
     0.48 & 
     5.08 & 
     5.37 & 
     10.45 \\
    \bottomrule
    \end{tabular}%
    }
\end{table*}

To examine whether our negotiation-based experiment aligns with commonly used LLM benchmarks such as MMLU, HumanEval, MATH, and GPQA \cite{docsbot_gpt4o_vs_deepseek, aimlapi_gemini_vs_gpt4o, evolution_ai_claude_gpt4o_gemini}, we present each model’s {benchmark scores} alongside the negotiation metrics (e.g., average sale price,  Win Rate, and persona influence) in Table~\ref{tab:benchmark-comparison}.

Our analysis shows that models with higher scores on existing benchmarks generally tend to exhibit strong performance in the negotiation game, demonstrating higher win rates and more favorable transaction prices. For instance, {claude-3-5-sonnet-20241022} and {gpt-4o}, both recognized as high-performing on MMLU and HumanEval, also achieved high win rates and robust strategic responses in actual negotiation tasks. This suggests that a sufficiently advanced level of {language comprehension and reasoning} can positively impact performance in complex, dialogue-driven environments such as our {negotiation game}.

Nevertheless, there are notable exceptions. For example, {gpt-4-turbo} attains relatively high scores on conventional benchmarks but manifests an extremely low win rate (approximately 0.08) when acting as a Seller. This discrepancy indicates that knowledge-based question answering and code interpretation benchmarks do not fully capture a model’s {social interaction} or {strategic dialogue} capabilities. Similarly, {gemini-1.5-flash}, which scores poorly on both conventional benchmarks and our negotiation test, reveals shortcomings not only in handling negotiation scenarios but also in broader reasoning and language skills.

Furthermore, when we use the total range metric to assess each model’s sensitivity to persona-driven emotional or behavioral imitation, we see that certain models (e.g., {claude-3-5-sonnet-20241022}) excel not only on standard benchmarks but also in reproducing a wide range of {social behaviors}, resulting in strong performance and high variability across different personas. In contrast, high-achieving models like {gpt-4o} and {gpt-4-turbo} demonstrate a narrower response to persona changes, suggesting that their negotiation styles and social strategies are less malleable.

In summary, although we observe a general {positive correlation} between standard benchmark scores and negotiation outcomes, there is not a perfect one-to-one relationship. Factors such as {social behavior imitation} and {strategic thinking} are insufficiently represented in traditional metrics. This finding underscores the value of negotiation simulations like {Buy and Sell} as a complementary benchmark for assessing a model’s {human-interaction} proficiency. 

\section{Conclusion}
This study quantitatively assessed how effectively Large Language Models can imitate human-like social behaviors and emotional expressions by conducting a {negotiation simulation} using the {Buy and Sell} game. Specifically, we (1) assigned different {personas} to each model to observe variations in negotiation outcomes, (2) analyzed various {Outcome Metrics} such as agreement success and final transaction price, (3) employed an XGBoost-based {Shap Value} analysis to quantify each persona’s contribution, and (4) compared our results to established benchmarks in order to explore correlations between LLMs’ {traditional language proficiency} and {strategic interactive capabilities}.

Our findings reveal that models scoring well on existing benchmarks generally also perform effectively in negotiation tasks, achieving higher win rates and more favorable transaction prices. However, this is not always the case; for instance, {gpt-4-turbo} excels in conventional metrics but underperforms in negotiation scenarios emphasizing {social and strategic contexts}. Such exceptions highlight the insufficiency of current benchmarks in evaluating {social behavior imitation} or {negotiation strategy} skills.

We also observe that persona assignments have a significant impact on negotiation outcomes. In particular, “aggressive/strategic” personas like {Competitive} and {Cunning} consistently produce higher win rates for both Buyers and Sellers. Meanwhile, personas such as {Altruistic} and {Cooperative} frequently lead to disadvantageous results across roles, a trend further corroborated by Shap Value analyses indicating their negative contribution. These interactions underscore that even the same LLM can exhibit dramatically different negotiation strategies and final deals based on its {social role} and {attitudinal prompts}.

Moreover, evaluating the magnitude of each persona’s influence via {Shap Values} shows that certain models (e.g., {claude-3-5-sonnet-20241022}) are highly responsive to persona changes, demonstrating a broad range of human-like emotional and social behaviors. In contrast, other models (e.g., {gpt-4-turbo}) maintain relatively consistent strategies regardless of the assigned persona, suggesting lower sensitivity to social or emotional cues. This finding supports the hypothesis that higher-performing LLMs can more flexibly and realistically replicate human behavioral and emotional dynamics.

In conclusion, this study contributes a new perspective on evaluating LLMs beyond their {language comprehension ability} by investigating how effectively they can enact {social and strategic interactions}. By simplifying and reproducing real-world elements such as {role allocation}, {asymmetric information}, and {persona-based emotion expression} within the negotiation game, we provide a method to quantify the aspects of {interpersonal aptitude} and {strategic decision-making} that are often overlooked in traditional benchmarks. Future research could explore multi-party negotiations, long-term trust building, emotional conflict resolution, or human-LLM collaborative experiments to more precisely gauge LLMs’ {human behavioral imitation} and {real-world applicability}.

\bibliographystyle{IEEEtran} % Use the IEEEtran BibTeX style
\bibliography{references.bib}

\vspace{12pt}
% \color{red}
% IEEE conference templates contain guidance text for composing 
% and formatting conference papers. Please ensure that all template text is 
% removed from your conference paper prior to submission to the conference. 
% Failure to remove the template text from your paper may result in your 
% paper not being published.

\end{document}